\documentclass[11pt,a4paper]{article}
\usepackage{naaclhlt2018}
\usepackage{times}
\usepackage{latexsym}
\usepackage{booktabs}
\usepackage{todonotes}
\usepackage{linguex}
 \usepackage{array,multirow,graphicx}
 
\usepackage{url}

\usepackage[usenames,dvipsnames]{pstricks}
\usepackage{epsfig}

\usepackage{enumitem} % allow removing enumerate left padding
\usepackage[hang,flushmargin]{footmisc} % remove footnote left padding

\aclfinalcopy % Uncomment this line for the final submission
\title{A Predictive Model for Notional Anaphora in English}

\author{Amir Zeldes \\
  Department of Linguistics \\Georgetown University \\
  Washington, DC, USA\\
  {\tt amir.zeldes@georgetown.edu} \\
  }

\date{}

\begin{document}
\maketitle
\begin{abstract}
Notional anaphors are pronouns which disagree with their antecedents' grammatical categories for notional reasons, such as plural to singular agreement in: ``the government ... they''. Since such cases are rare and conflict with evidence from strictly agreeing cases (``the government ... it''), they present a substantial challenge to both coreference resolution and referring expression generation. Using the OntoNotes corpus, this paper takes an ensemble approach to predicting English notional anaphora in context on the basis of the largest empirical data to date. In addition to state of the art prediction accuracy, the results suggest that theoretical approaches positing a plural construal at the antecedent's utterance are insufficient, and that circumstances at the anaphor's utterance location, as well as global factors such as genre, have a strong effect on the choice of referring expression.
\end{abstract}

\section{Introduction}

In notional agreement, nouns which ostensibly belong to one agreement category are referred back to using a different category, as in \ref{ex_quirk} \cite{QuirkGreenbaumLeechEtAl1985}, with singular/plural verb and pronoun.

\ex. [The government] has/have voted and [it] has/[they] have announced the decision\label{ex_quirk}

Although examples such as \ref{ex_quirk} are often taken to represent a single phenomenon, subject-verb (SV) agreement and pronoun number represent distinct agreement phenomena and can disagree in some cases, as shown in \ref{CNN} and \ref{hospital}, taken from the OntoNotes corpus \cite{HovyMarcusPalmerEtAl2006}.

\ex. [CNN] \textbf{is} my wire service; [they]\textbf{'re} on top of everything.\label{CNN}

\ex. [One hospital] in Ramallah \textbf{tells} us [they] \textbf{have} treated seven people\label{hospital}

While previous studies have focused on SV agreement (\citealt{Dikken2001}, \citealt{Depraetere2003}, \citealt{InsuaMartinez2003}), there have been few corpus studies of notional pronouns, due at least in part to the lack of sizable corpora reliably annotated for coreference, and the low accuracy of automatic systems on difficult cases. In this paper we take advantage of the OntoNotes corpus, the largest corpus manually annotated for coreference in English (about 1.59 million tokens with coreference annotations), to build a predictive model of the phenomenon, which can be used for both coreference resolution and referring expression generation (see \citealt{KrahmerDeemter2012} for an overview).

\section{Previous work}\label{prev_work}

Theoretical linguistic discussions have focused on SV agreement, especially in expletive constructions (ECs, \citealt{Sobin1997}; i.e. `there is' vs. `there are'). \citet{Reid1991} discusses SV agreement and notional pronouns, and posits reference to persons as facilitating plural pronouns, as in \ref{couple} and \ref{court}, where a relative `who' forces a +PERSON reading. 

\ex. And this fall [the couple] expects [its] first child.\label{couple}

\ex. A Florida court ruled against [a Pennsylvania couple] \textbf{who} contend May's 10-year-old daughter is actually [their] child.\label{court}

This suggests that inferred entity type may be a relevant predictor of notional anaphora. Other theoretical papers suggest a formal analysis with empty pronoun heads bearing a plural feature, e.g. the analysis in \ref{empty_pro} from \citet{Dikken2001} (see also \citealt{Sauerland2003} for a similar analysis). 

\ex. [\textsubscript{DP1} pro\textsubscript{[+pl]} [\textsubscript{DP2} the committee\textsubscript{[-pl]}]] are ... \label{empty_pro}

This suggests that speakers decide on the notional agreement category already at the point of uttering the antecedent. However psycholinguistic studies have shown effects localized to the point of uttering the anaphor, due to processing constrains (see \citealt{EberhardCuttingBock2005}, \citealt{WagersLauPhillips2009}, \citealt{Staub2009}). We hypothesize that processing constraints may make it difficult for speakers to remember the exact expression used for the antecedent after a long distance from the first point of utterance, and therefore consider some length and distance-based metrics as features below (see Section \ref{features}).

Corpus-based studies have shown that notional anaphora likelihood varies by modality (more often in speech), variety of English (more often in UK English) and genre (see \citealt{QuirkGreenbaumLeechEtAl1985}: 758, \citealt{LeechSvartvik2002}: 201, \citealt{Levin2001}). \citet{Depraetere2003} explored the idea that verb semantics influence agreement choice, especially whether verbs imply decomposition or categorization of the unit (e.g. \textit{consist of, be gathered, scatter}), or signify differentiation within a set (e.g. \textit{disagree, quarrel}). \citet{Annala2008} provides a detailed corpus study of nine nouns in the written part of the British National Corpus (\url{http://www.natcorp.ox.ac.uk/}) and the Corpus of Late Modern English Texts (CLMETEV, \url{http://www.helsinki.fi/varieng/CoRD/corpora/CLMETEV/}). The study found tense to be relevant for the nine nouns, with past tense of `be' being particularly susceptible to triggering plural agreement, while for nouns which generally prefer plural, singular agreement appeared more often in the present. Taken together, these studies suggest that tense and verb classes may be relevant features, as well as indicating the importance of some conventional usage effects. The latter are also backed by psycholinguistic evidence that speakers process notional anaphora more quickly than strict agreement in contexts that are biased towards the non-agreeing plural \cite{Gernsbacher1986}.

\section{Experimental setup}\label{setup}

\subsection{Types of cases included}\label{ana_types}

In this paper we focus exclusively on plural pronouns referring back to singular headed phrases, but the exact nature of cases included requires some decisions. Since the number for second person pronouns (\textit{you}, \textit{your}, etc.) is ambiguous, we omit all second person cases. First person cases are rare but possible, especially in reference to organizations, as in \ref{SEC}, taken from OntoNotes.

\ex. Bear Stearns stepped out of line in the view of [the SEC]\textsubscript{i} ... [we]\textsubscript{i}'re deadly serious about bringing reform into the marketplace\label{SEC}

Some of the same lexical heads can appear with both singular and plural first person reference, either for metonymical reasons (``when a country says `I/we'...'') or by coincidental homonymy.\footnote{For example one speaker in a forum discussion in the corpus has the user name `A Very Ordinary Native Country', leading to coreference with the pronoun `I'.} These cases are therefore all included whenever a relevant NP is annotated as coreferent in OntoNotes. 

Three main types of plural reference to singular antecedents can be distinguished in our data (see Section \ref{data} for some statistics): the most common, which will be referred to as Type I, is reference to complex/distributive entities (so called `committee' nouns) seen e.g. in \ref{CNN}. These are distinct from Type II, which has bleached quantity noun heads, (e.g. `a number of X' or `a majority of X') which may sometimes be referenced as a plurality, as in \ref{majority_pl}, and sometimes as a unit, as in \ref{majority_sg}.

\ex. [the vast silent \textbf{majority} of these Moslems] are not part of the terror and the incitement , but [they] also do not stand up political leaders\label{majority_pl}

\ex. [the vaunted Republican \textbf{majority}] is just not now nor has [it] ever been ready for prime time governing\label{majority_sg}

A third type (Type III) occurs in cases such as \ref{epicene}, denoting unspecified gender (these are sometimes called generic or epicene pronouns; see also \citealt{HuddlestonPullum2002}:493-494, \citealt{Curzan2003}). This construction has been gaining popularity \cite{Paterson2011}, and has recently been approved by the 2017 Associated Press Stylebook as standard (\url{https://www.apstylebook.com/}).

\ex. I'll go and talk to [the person here] cause [they] get cheap tickets\label{epicene}

Although this type of agreement is semantically and pragmatically very different from the other two types above, it must be addressed in this paper for several reasons. Firstly, if we want to be able to predict pronoun form for computational applications such as coreference resolution or natural language generation, then such cases should be covered in some way. Secondly, there are cases in which either a computer, or in some cases even a human would find it difficult or impossible to be sure of the class that a case falls under, as shown in \ref{enemy} and \ref{publisher}, both real examples from OntoNotes.

\ex. [The enemy] attacked several times, and each time [they] were repelled \label{enemy}

\ex. [a publisher] is interested in my personal ad book ... I looked [them] up \label{publisher}

While in \ref{enemy} it may seem  unlikely that use of `they' is meant to obscure gender, this reading cannot be ruled out, especially by automatic analysis. In \ref{publisher}, it is possible to get either reading: either the `publisher' is a company, and therefore plural (Type I; but notice singular `is' as a verb), or the speaker spoke with the director of a publishing house, disregarding that person's gender (Type III). Note that in singular agreement, these would result in saying `she' or `he' versus `it', as in \ref{desmoines} (also from OntoNotes).

\ex. [the Des Moines-based publisher] said [it] created a new Custom Marketing \label{desmoines}

Additionally, there are Type III cases in which plural pronoun agreement for singular-like reference is not motivated by gender constraints, e.g. \ref{nobody}. 

\ex. [Nobody] is going to like Bolton a year from now, are [they]? 
\label{nobody}

Due to these complications, we include all cases of plural anaphora annotated as coreferent with singular NPs, though we will re-examine these types in the data in analyzing the results. 

\subsection{Data}\label{data}

The data for the present study comes from the OntoNotes corpus \cite{HovyMarcusPalmerEtAl2006}, Version 5, the largest existing corpus with coreference annotations. OntoNotes contains gold POS tags and syntactic constituent parses, as well as coreference resolution for pronominal anaphora and definite or proper noun NPs (but not for indefinites, see below), and named entity annotations for proper nouns. The coreference annotated portion of the corpus contains 1.59 million tokens from multiple genres, presented in Table \ref{onto_texttypes}. 

\begin{table}[h!]
  \begin{center}
    \caption{Coarse text types in OntoNotes}
    \label{onto_texttypes}
    \begin{tabular}{l|r|l|r} 
       \multicolumn{2}{c|}{\textbf{Spoken}} & \multicolumn{2}{c}{\textbf{Written}}\\
      \hline
      bc.conv & 137,223 & news & 68,6455 \\
      bc.news & 244,425 & bible & 243,040 \\
      phone & 110,132 & trans. & 98,143 \\
       &  & web & 71467 \\
      \hline
       \textbf{total} & 491,780 & \textbf{total} & 1,099,105 \\
      \hline
       \multicolumn{4}{c}{\textbf{total} 1,590,885}
    \end{tabular}
  \end{center}
\end{table}

Written data constitutes the large bulk of material, primarily from newswire (Wall Street Journal data), as well as some data from the Web and the New Testament, and some translations of news and online discussions in Arabic and Chinese. The translated data has been placed in its own category: it behaves more conservatively in preferring strict agreement than non-translated language (see Section \ref{individual_predictors}), perhaps due to translators' editorial practices. The spoken data comes primarily from television broadcasts, including dialogue data from MSNBC, Phoenix and other broadcast sources (bc.conv), or news, from CNN, ABC and others (bc.news), as well as phone conversations.

The relevant cases from the corpus for the present study were extracted by finding all lexical NPs headed by singulars (tagged NNP or NN) whose phrases are referred back to by an immediate antecedent (the next mention) which is a first or third person pronoun, then filtering to keep only those singular NPs headed by a token which is attested as taking plural agreement somewhere in the corpus, but also including its attestation with singular pronouns. In other words, this study makes no \textit{a priori} interpretation of anaphora as notional in isolation: all and only items actually attested in both forms are considered. 

These selection criteria, followed by manual filtering for errors, led to the extraction of 3,488 anaphor-antecedent pairs, of which 1,209 exhibited notional agreement (34.6\%), including a subset of 207 cases (5.9\% of the data) which were unambiguously identifiable as Type III, gender neutral plural pronouns. 

OntoNotes contains 17,263 direct anaphoric links to a singular NP, meaning we can estimate the frequency of all agreement types addressed here at a not insubstantial 7\% of pronominal reference to a singular lexical NP antecedent, with gender neutral type III at about 1.2\% and Types I-II covering 5.8\% of the total corpus. 

As a test data set, we reserve a random 10\% of the data, amounting to 349 cases, stratified to include approximately the same proportions of genres, as well as notional vs. strict agreement cases. This stratification is important in order to test the classifier in Section \ref{ensemble} using realistically distributed data. 

\subsection{Feature extraction}\label{features}

To predict the occurrence of notional anaphora we will use a range of categorical features indicated to be relevant in previous studies (see Section \ref{prev_work}): POS tags and dependency functions for the anaphor, antecedent and their governing token, entity types, genre/modality, and definiteness/previous mention of the antecedent. These features indirectly give us access to tense, grammatical constructions and some measures of salience (especially subjecthood and repeated mention). Additionally, we will consider a number of numerical features which may be relevant from a processing perspective, such as the distance in tokens between the anaphor and antecedent, length in characters and tokens for the antecedent NP, document token count, and the positions of the expressions in the document, expressed as a percentage of document length (e.g. an antecedent may begin at the 75\textsuperscript{th} percentile of document token count). Most of these features can be extracted from the data automatically.

A limitation of using OntoNotes is that many antecedents of pronominal anaphora are not named entities (unnamed `committees', etc.), meaning we do not have gold entity types for all NPs. In order to overcome this problem and expand the range of features available in this study, the entire corpus was annotated automatically for non-named entities using xrenner, a non-named entity recognizer \cite{ZeldesZhang2016}. 

A second problem is that the coreference annotation guidelines for OntoNotes preclude antecedents for indefinite NPs, meaning cases such as \ref{indef_coref} are marked as multiple entities (\citealt{BBNTechnologies2007}:4). 

\ex. [Parents]\textsubscript{x} should be involved with their children's education at home, not in school. [They]\textsubscript{x} should see to it ... [Parents]\textsubscript{y} are too likely to blame schools for the educational limitations of [their]\textsubscript{y} children.\label{indef_coref}

The second instance of `parents' is regarded as a separate, `discourse new' entity. This will be relevant for using previous mention of the antecedent as a feature: we can only detect previous mention of the antecedent if it is annotated, and this will never be the case for indefinites. 

In order to assess the influence of grammantical function and semantic classes of verbs governing either the anaphor or the antecedent, the syntax trees in the corpus were converted to a dependency representation using Stanford CoreNLP \cite{ManningEtAl2014}, allowing for a simpler use of dependency functions as a predictor. This also allows us to identify the governing verb (or noun etc.) for each mention. Governing verbs were then tagged automatically using VerbNet classes (Kipper et al. 2006), which give rough classes based on semantics and alternation behaviors in English, such as ALLOW for verbs like \{\textit{allow, permit, ...}\} or HELP: \{\textit{aid, assist, ...}\}, etc. 

Because some verb classes are small or rare, potentially leading to very sparsely attested feature values, classes attested fewer than 60 times were collapsed into a class OTHER (for example VerbNet class 22.2, AMALGAMATE). Verbs attested in multiple classes were always given the majority class, for example the verb \textit{say} appears both in VerbNet class 37.7 SAY and class 78, INDICATE, but was always classified as the more common SAY.  Finally, some similar classes were collapsed in order to avoid replacement by OTHER, such as LONG + WANT + WISH, which were fused into a class DESIRE. Nominal governors (e.g. for possessive NPs, whose governor is the possessor NP) were classified by their NER entity class or non-named class predicted by the entity recognizer.

\section{Results}\label{results}

\subsection{Predictive ensemble model}\label{ensemble}

In this section we construct a model to predict, given properties of a singular antecedent NP from a lexeme known to exhibit notional agreement, and properties of the position of the anaphor referring back to it, whether or not the pronoun will in fact be plural. Considering the highly contextual nature of notional anaphora, we would ideally want to use the entire sequence of text before and after each of the entity mentions to predict the choice of pronoun, for example using a Recurrent Neural Network. However, despite being the largest available dataset for English, the amount of gold standard examples we have (less than 4,000) makes a Deep Learning approach problematic. We therefore train an ensemble of decision trees on the features presented in Section \ref{setup}, more specifically using the Extra Trees algorithm \cite{GeurtsErnstWehenkel2006}, which outperforms the standard Random Forest algorithm and linear models on our data. 

\begin{figure}[!ht]
\centering
\includegraphics[width=0.5\textwidth]{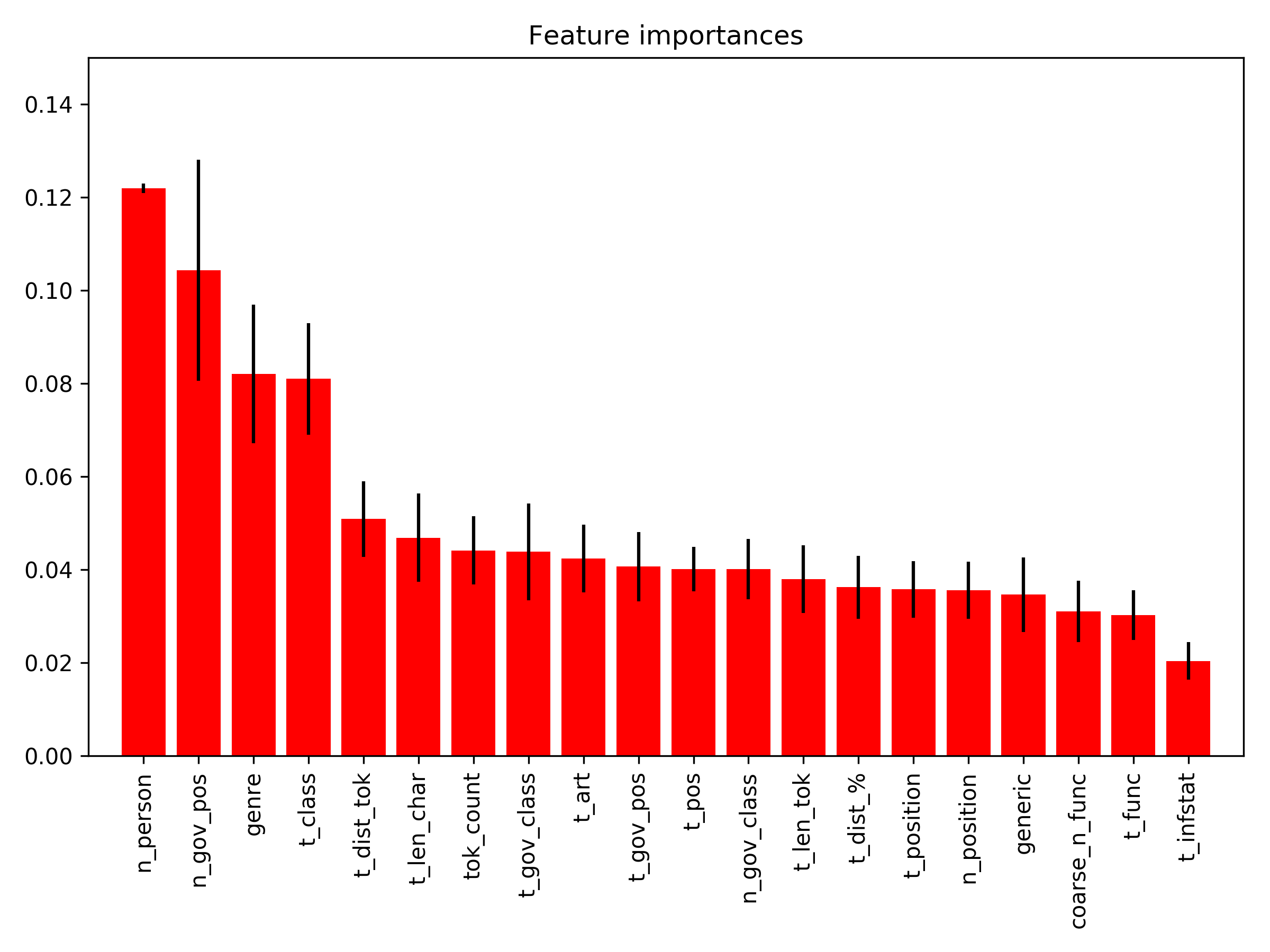}
\caption{Variable importances for the classifier. Features beginning with n\_ apply to the anaphor, and features with t\_ to the antecedent.}
\label{importances}
\end{figure}

Using a grid search with 5 fold cross validation on the training data, the optimal hyper-parameters for the classifier were found, leading to the use of 300 trees with unlimited depth, limited to the default number of features in the scikit-learn implementation, which is the square root of the number of features rounded up. The best performance was achieved using the 20 features outlined in Figure \ref{importances}, meaning that each tree receives 5 features to work with, thereby reducing the chance of overfitting training data. The classifier achieves a classification accuracy of 86.81\% in predicting the correct form in the test set, an improvement of over 20\% above the majority baseline of always guessing `strict agreement' (65.6\% accuracy).

To evaluate the importance of features in Figure \ref{importances} we use the Gini index of purity achieved at splits using each respective feature across all trees. Error bars indicate the standard deviation from the average importance across all trees in the ensemble. A Gini index of 0 means complete homogeneity (for our task, a 50-50 split on both sides), whereas 1 would mean perfect separation based on that feature. In addition to features discussed above, a feature `generic' was introduced for phrases such as `anyone', `someone', `somebody', etc. which behave differently from other PERSON entities, as well as a feature `t\_art' coding the antecedent's article as definite, indefinite, demonstrative, or none.

The most important feature is 1st vs. 3rd person anaphor (`n\_person'), as these are rather different situations: 1st person cases occur mainly with individuals speaking for aforementioned organizations, introduced as proper nouns (e.g. `the SEC ... we' in \ref{SEC}). Next is the POS tag of the anaphor's governor, which includes information about tense and can work in conjunction with verbs' semantic classes and grammatical functions (cf. \citealt{Depraetere2003}, \citealt{Annala2008}). Genre is surprisingly important in third place (cf. \citealt{Levin2001}), indicating that settings licensing notional anaphora are genre specific. Replacing genre with a more coarse grained spoken/written variable degrades accuracy. Genre is closely followed by the semantic class of the antecedent, i.e. the entity in question, which is clearly relevant (+/-PERSON and more, see Section \ref{individual_predictors} for details).

Subsequent variables are less important, including distance, length and position in the document. Though both are helpful, using the article form (`t\_art') is more important than the information status or previous mention (`t\_infstat') based on antecedents to the antecedent (keeping in mind limitations of the coreference annotations, cf. Section \ref{data}). Grammatical functions are helpful, but less so than other features. 

Looking at the actual classifications obtained by the classifier produces the confusion matrix in Table \ref{confusion}. The matrix makes it clear that the classifier is very good at avoiding errors against the majority class: it almost never guesses `notional' when it shouldn't. Conversely, about 1/3 of actual notional cases are misclassified, predicted to be `strict'. Among the erroneous cases, only 6 belong to Type III (about 15\% of errors) , showing that the classifier largely handles this type quite well next to the other types, since Type III covers about 20\% of plural-to-singular agreement cases. 

\begin{table}[!ht]
\small
  \begin{center}
    \caption{Confusion matrix for test data classification}
    \label{confusion}
\begin{tabular}{l|l|c|c|c}
\multicolumn{2}{c}{}&\multicolumn{2}{c}{Predicted}&\\
\cline{3-4}
\multicolumn{2}{c|}{}&Sg&Pl&\multicolumn{1}{c}{Total}\\
\cline{2-4}
\multirow{2}{*}{Actual}& Sg & 222 & 39 & 261\\
\cline{2-4}
& Pl & 7 & 81 & 88\\
\cline{2-4}
\multicolumn{1}{c}{} & \multicolumn{1}{c}{Total} & \multicolumn{1}{c}{229} & \multicolumn{1}{c}{120} & \multicolumn{1}{c}{349}\\
\end{tabular}
\end{center}
\end{table}

\subsection{Analysis of predictors}\label{individual_predictors}

\begin{figure}[hbt]
\centering
\includegraphics[width=0.5\textwidth]{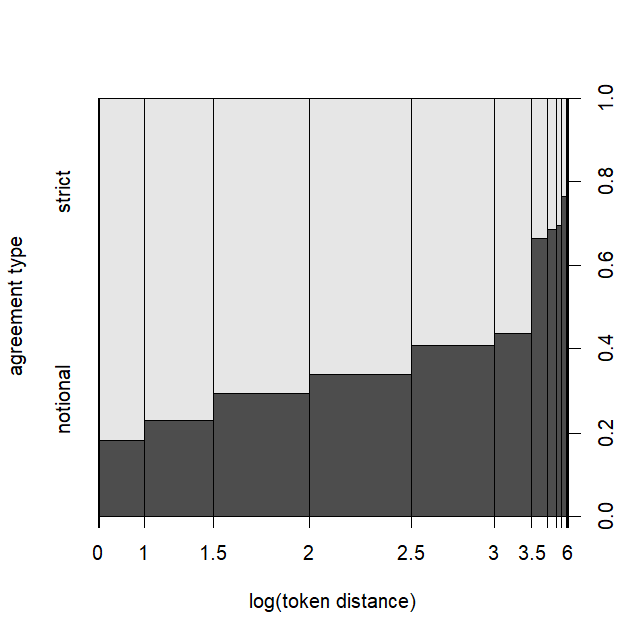}
\caption{Log token distance between anaphor and antecedent.}
\label{distance}
\end{figure}

\begin{figure}[hbt]
\centering
\includegraphics[width=0.5\textwidth]{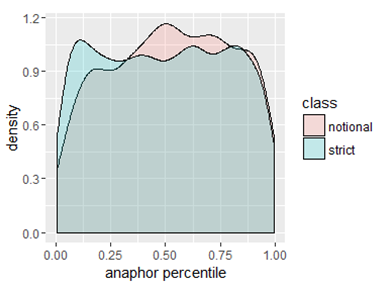}
\caption{Position of anaphor as percentile of document length in tokens.}
\label{position}
\end{figure}

To understand why the features used in the previous section are helpful we analyze the distribution of notional anaphors for several non-obvious predictors individually. Beginning with processing factors, we can consider the effect of distance between anaphor and antecedent and position in the document, shown in Figures \ref{distance} and \ref{position}.

In Figure \ref{distance}, token distance is shown in log-scale, as greater distances are attested sparsely, and the breadth of each column in the spine plot corresponds to the amount of data it is based on. It is easy to see the perfectly monotonic rise in the proportion of notional agreement, beginning with under 20\% at a log-distance of \textasciitilde1, all the way to over 50\% at log-distances of \textasciitilde3.5 or higher (approximately 33 tokens and above).

Figure \ref{position} shows why position in the document matters: there is a slightly higher frequency of notional agreement after the halfway point of documents. This can be related to a speaker fatigue effect (speakers/writers become less constrained and exhibit less strict agreement as the document goes on), or due to editorial fatigue in written data (editors correct notional agreement, but notice it less frequently further in the document). However while we would only expect an editorial motivation to affect written data, the effect is found in both spoken and written documents, meaning a possible speaker fatigue effect cannot be discounted.

Next we can consider the effect of genre, and expectations that speech promotes notional agreement. This is confirmed in Table \ref{genres}. However we note that individual genres do behave differently: data from the Web is closer to spoken language. The most restrictive genre in avoiding notional agreement is translations. Both of these facts may reflect a combination of modality, genre and editorial practice effects. However the strong differences suggest that genre is likely crucial to any model attempting to predict this phenomenon.

\begin{table}[h!]
  \begin{center}
    \caption{Agreement patterns across genres}
    \label{genres}
    \begin{tabular}{l|r r r} 
       \textbf{genre} & \multicolumn{3}{c}{\textbf{agreement}} \\
       \textit{written} & \textit{notional} & \textit{strict} & \textit{\% notional} \\
      \hline
      bible & 169 & 487 & 25.76 \\
      newswire & 344 & 843 & 28.98 \\
      translations & 55 & 210 & 20.75 \\
      web & 48 & 71 & 40.33 \\
      \textbf{total written} & 616 & 1611 & 27.66 \\
      \hline
       \textit{spoken} & \textit{notional} & \textit{strict} & \textit{\% notional} \\
      \hline
      bc.conv & 237 & 201 & 54.11 \\
      bc.news & 296 & 378 & 43.91 \\
      phone & 60 & 89 & 40.26 \\
      \textbf{total spoken} & 593 & 668 & 47.02 \\
    \end{tabular}
  \end{center}
\end{table}

\begin{figure}[hbt]
\centering
\includegraphics[width=0.5\textwidth]{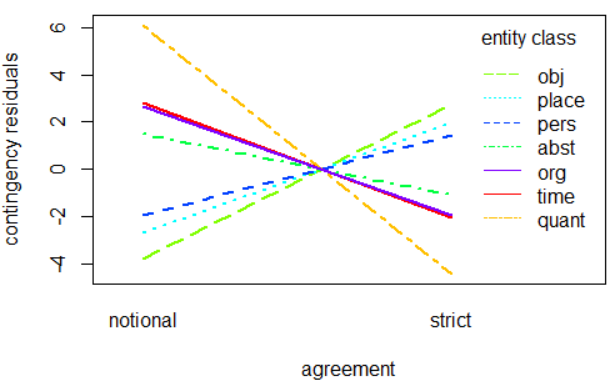}
\caption{Chi square residuals for notional agreement by entity type. The legend is ordered by strictness.}
\label{entity_resids}
\end{figure}

Moving on to grammatical and semantic factors, we consider the entity type of the referring expression in Figure \ref{entity_resids}. The plot shows the chi square residuals for the association of each entity type with the two agreement types. Lines sloping top-right to bottom-left correspond to entity types preferring strict agreement (OBJECT, PLACE, PERSON), while top-left to bottom-right slopes correspond to types preferring notional agreement (QUANTITY, TIME, ORGANIZATION). 

The result that PERSON somewhat prefers strict agreement is surprising given the expectation that agentive, human-associated predicates have an effect promoting notional agreement \cite{Depraetere2003}. This is because many of those predicates were most often associated in our data with an ORGANIZATION telling, having or wanting to do something, and then being construed as a group of humans. This leads to the notional preference of the ORGANIZATION class. NPs actually classified as PERSON often included heads such as the very common \textit{family} (mostly singular agreement), or potential Type III nouns which often take explicit gender (e.g. gender-specific \textit{`baby ... her/his'}). Less surprising is the association of QUANTITY and TIME with notional agreement, covering cases such as \textit{`a third of ... they'}, and counted time units in Type II phrases such as \textit{`a couple of (minutes/hours)'}.

\begin{figure*}[hbt]
\centering
\includegraphics[width=\textwidth]{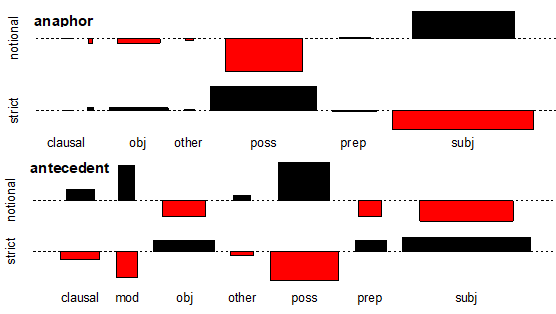}
\caption{Association of pronoun choice and dependency functions of the anaphor and antecedent (top: anaphor; bottom: antecedent). The category `clausal' collapses the labels `csubj', `ccomp' and `advcl'.}
\label{deprel}
\end{figure*}

Looking at the distribution of grammatical forms and functions, Table \ref{n_parent_pos} shows imbalances based on the POS tag of the token governing the anaphor, and Figure \ref{deprel} shows an association plot between dependency functions\footnote{Two versions of the function labels were tested: coarse labels as used in Figure \ref{deprel} (e.g. `subj', `clausal') and all available labels in the Stanford CoreNLP basic label set (distinguishing active `nsubj' and passive `nsubjpass', different types of clauses, etc.). The classifier works best with coarse labels for the anaphor's function but fine grained ones for the antecedent.} and agreement patterns (rare POS and dependency labels have been omitted for clarity). 

\begin{table}[!ht]
  \begin{center}
    \caption{Agreement by anaphor governor POS}
    \label{n_parent_pos}
    \begin{tabular}{l|r r r} 
 & notional & strict & \% notional \\
 \hline
VBG & 112 & 94 & 54.36 \\
VBpres\footnotemark & 218 & 255 & 46.08 \\
VB & 244 & 291 & 45.61 \\
JJ & 48 & 82 & 36.92 \\
VBD & 183 & 313 & 36.89 \\
IN & 65 & 117 & 35.71 \\
VBN & 81 & 163 & 33.19 \\
NNP & 8 & 18 & 30.76 \\
NN & 141 & 645 & 17.93 \\
\end{tabular}
\end{center}
\end{table}

The table confirms the observations by \citet{Annala2008} that present tense favors plural agreement more than past tense (VBD/VBN), but also reveals that nominal governors (NNP and more so NN, primarily possessed nouns of the entity in question), also promote singular agreement. This is echoed in the association plot in Figure \ref{deprel}. Possessive anaphors (`poss') prefer strict agreement and anaphoric subjects promote plural agreement, while the opposite is true for antecedents: if the antecedent is a subject, it is more likely to be realized later as a singular, and the opposite if it is a possessive. 

\footnotetext{The tags VBP and VBZ have been collapsed into VBpres, since they trivially imply whether the anaphor was singular or plural.}

It is possible that the increased salience of subjects adds to speakers' tendency to refer back to them in keeping with the morphological number of the previous mention, while a late mention as a subject allows the salient anaphor position to select a disagreeing form more easily, without depending on previous mentions. Investigating this hypothesis further may require psycholinguistic data.

\section{Conclusion}

One of the fundamental challenges of notional agreement  is the apparent unpredictability shown often in previous studies: the same nouns can appear under seemingly similar conditions with both types of agreement. The ensemble classifier presented here shows that despite this unpredictability, comparatively good predictions can be made on unseen data, with an accuracy of 86.81\%, substantially improving on a baseline of 65.6\%.\footnote{An anonymous reviewer has asked to what extent state of the art coreference resolution systems also err on notional cases  in general and the cases targeted here in particular: this is an interesting question which probably depends on the system, but it certainly seems possible that some architectures could benefit from notional agreement probability estimation, similarly to preprocessors predicting singleton status \cite{RecasensEtAl2013} or other special constructions (e.g. anaphoric `one' in English, \citealt{RecasensHuRhinehart2016}).} 

The classifier showed a good ability to recognize the majority class, but also learned to be `cautious', guessing `strict' in 1/3 of notional cases. A possible interpretation of this result is that for ambiguous cases, in which either form could be acceptable, the classifier chooses the safer majority class. In many such misclassified cases it seems likely that speakers would accept either variant, as in \ref{comsat}, which the classifier gets wrong:

\ex. [Comsat Video, which distributes pay-per-view programs to hotel rooms], plans to add Nuggets games to [their] offerings\label{comsat}

In this example, multiple signals suggest strict agreement, including an aforementioned, subject antecedent, and short distance to the 3rd person possessive anaphor. Based on features from the training data, it is a fair example of the environment of a `strict' case; at the same time, it seems likely that speakers would accept a version with `its', and it is not difficult to find similar examples, with similar distances, syntax and governing items, as in \ref{school}.\footnote{An anonymous reviewer has suggested that checking human acceptability of such deviating cases would be an interesting follow up study, and we certainly agree.} 

\ex. Ultimately, Lewis said, [her school] added African-American history to [its] offerings\label{school}

Another aspect worth considering is the feature space used here, and some possible alternatives. Among the features tested but ultimately rejected in this study, we examined the presence of relative clause markers as suggested by \citet{Reid1991}, as well as some alternative semantic representations for governing verb semantics. For relative clauses, the importance of cases with `who' as in example \ref{couple} turned out not to be useful in practice, despite the presence of well over 200 relative clauses in the data and over 150 with `who(m)'. It can be suspected that relative pronouns modifying the antecedent at the point it is mentioned have less interactions with anaphors, which can appear much later in the text, than with immediate subject-verb agreement cases which motivated the observation in \citet{Reid1991}.

For encoding verb semantics, the choice of VerbNet categories and the lack of disambiguation for ambiguous cases are both far from optimal. VerbNet classes do not necessarily map well onto verb groups' preferences for notional agreement. It seems likely that other thematic, cluster-based or vector space-based methods of classifying verb semantics could be helpful for the present task. To this end we tested using semantic classes as assigned by the UCREL Semantic Analysis System (USAS, \citealt{RaysonArcherPiaoEtAl2004}), which performed worse than VerbNet. Some VerbNet classes are mirrored in the USAS classes (e.g. communication verbs, the USAS coarse domain Q, or sub-classes in domain Q2); however in many cases it is possible that, by being much more specific (classes such as `science and technology' in USAS), content domain classes encourage the classifier to memorize specific training instances, which do not generalize well. Ideally, a flexible semantic representation such as trainable embeddings would likely be helpful, but would require training on an external dataset beyond the notional agreement pairs, which only amount to a few thousand examples.

For future work, we can point out that while the classifier achieved overall good accuracy above chance, there is substantial room for improvement, and more features could be considered. These include phonological features (e.g. phonotactics around anaphors, metrical factors), morphological features (affixes, types of compounding), semantic features (more directly targeting predicates with distributive readings) and further context cues such as modifiers (adjectives, adverbs) and other words in the context not directly governing or governed by the noun in question. For NLP and NLG applications, it would be most useful to consider those variables for which we can build automatic taggers or generated contexts in real-time. At the same time, it will probably remain impossible to achieve perfect accuracy: it is expected that, as with many high level alternations, some element of inter- and even intra-speaker variation, as well as speakers' communicative intentions, will always create a certain degree of unpredictability in settings which are otherwise comparable. 

%\bibliography{naaclhlt2018}
\bibliography{notional.bib}
\bibliographystyle{acl_natbib}

\end{document}